\documentclass{article}

\usepackage{authblk}
\usepackage{arxiv}
\usepackage{amsmath}
\usepackage{float}
\usepackage{multirow}
\usepackage{algpseudocode}
\usepackage{wrapfig}
\usepackage{algorithm}
\usepackage[utf8]{inputenc} 
\usepackage[T1]{fontenc}    
\usepackage{hyperref}       
\usepackage{url}            
\usepackage{booktabs}       
\usepackage{amsfonts}       
\usepackage{nicefrac}       
\usepackage{microtype}      
\usepackage{lipsum}		
\usepackage{graphicx}
\usepackage{natbib}
\usepackage{doi}

\title{A Geometric Algorithm for Blood Vessel Reconstruction from Skeletal Representation}

\author[1, 2]{
    Guoqing Zhang\thanks{wlsdzyzl@gmail.com} 
    }
\author[1]{
    Yang Li
    }

\affil[1]{Tsinghua Berkeley Shenzhen Institute (TBSI) , Tsinghua University}
\affil[2]{Peng Cheng Laboratory}

\date{}

\renewcommand{\headeright}{}
\renewcommand{\undertitle}{}

\hypersetup{
pdftitle={A Geometric Algorithm for Blood Vessel Reconstruction from Skeletal Representation},
pdfsubject={q-bio.NC, q-bio.QM},
pdfauthor={David S.~Hippocampus, Elias D.~Striatum},
pdfkeywords={},
}

\begin{document}
\maketitle
\begin{abstract}
We introduce a novel approach for the reconstruction of tubular shapes from skeletal representations. Our method processes all skeletal points as a whole, eliminating the need for splitting the input structure into multiple segments. We represent the tubular shape as a truncated signed distance function (TSDF) in a voxel hashing manner, in which the signed distance between a voxel center and the object is computed through a simple geometric algorithm. Our method does not involve any surface sampling scheme or solving large matrix equations, and therefore is a faster and more elegant solution for tubular shape reconstruction compared to other approaches. Extensive experiments demonstrate the efficiency and effectiveness of the proposed method. Code is available at \href{https://github.com/wlsdzyzl/Dragon}{https://github.com/wlsdzyzl/Dragon}.
\end{abstract}


\section{Introduction}
Shape reconstruction is a popular topic and has garnered significant attention from researchers due to its wide range of applications in computer graphics and computer vision. Existing methods focus on reconstructing object or scene from a point cloud or a sequence of image. Point cloud-based methods utilize a continuous high-dimensional function to fit the surface \cite{carr2003smooth, kazhdan2006poisson, hou2022iterative}. These methods usually involve solving large linear equations and heavily rely on the quality of the point cloud and accurate estimation of surface normals. Image-based approaches iteratively fuse input frames into a distance or radiance field \cite{niessner2013real, han2018flashfusion, mildenhall2021nerf, dong2018psdf}. Accurate camera pose estimation is necessary to generate a consistent radiance field. The above methods usually require the use of cameras or radars to capture the surface of the object, and have the ability to perform high-quality reconstruction of the geometry and texture details. However, they struggle to generate complete and topologically correct models due to scan occlusion. Post-processing algorithms such as point cloud completion \cite{huang2020pf, wen2020point, kraevoy2005template, dai2017shape} also have difficulty dealing with complex topological structures. As a consequence, these methods are not suitable for accurate shape reconstruction of tubular structures, particularly complex vascular trees, which are characterized by their intricate topology. 

Skeletons serve as compact representations of primitive structures with identical topology \cite{choi1997mathematical}, and are inherently well-suited for effectively representing tubular structures.  Skeletonization of tubular structures has undergone extensive research and demonstrated its utility across various applications \cite{zwettler2008accelerated, wang2020deep, shit2021cldice}. However, recovering original shape from skeletal representation has been less studied. In practice, obtaining the skeleton or centerline of an object is much easier compared to acquiring the complete shape of the object. For example, annotating the centerline of coronary artery in CCTA requires much less labor compared to annotating the entire vessel \cite{metz20083d,schaap2009standardized}. Shape reconstruction of centerlines can help generate binary mask volume and further improve the performance of segmentation model.

This study presents a novel approach for reconstructing tubular shapes from skeletal representations. In particular, each skeletal point is considered as a slice and the recovered shape is considered as a linear interpolation of these slices. Based on this simple assumption, we propose a pure geometric algorithm to compute the signed distance between any given location and the object. The space is represented as sparse voxels whose TSDF values are efficiently computed in parallel and stored in a spatial hashing table. Finally, we use a multi-threaded marching cubes algorithm to extract the isosurface as triangle meshes. In our approach, all skeletal points are processed simultaneously, thereby obviating the requirement for splitting the object into multiple segments. In addition, the proposed method does not involve any surface sampling scheme or solving large matrix equations, resulting in a faster and more elegant solution of tubular shape reconstruction. 
\section{Related Works}
\label{related_works}
\noindent\textbf{Tubular structure and skeleton representation:} Analysis of tubular structure plays an important role in many fields of research ranging from civil engineering to medical imaging. Various works \cite{xing2001numerical, hu2019topology, luo2021design, gharleghi2022towards} are proposed for designing and segmenting tubular structures, for which the topology is their most important characteristic. Skeletons are concise representations of original structures with identical topology \cite{choi1997mathematical} and are inherently well-adapted for efficiently depicting tubular structures. Sufficient research focus has been directed towards the skeleton extraction and application. \cite{huang2013l1, qin2019mass} extract curve skeleton from point clouds through medial axis transforms and mass-driven optimal transports. \cite{lin2021point2skeleton} represents skeleton points as weighted summations of surface points and use a deep network to estimate the weights. \cite{shit2021cldice, Zhang_2023_BMVC} introduce skeletonization in tubular structure segmentation and demonstrate its topology-preserving capability.  

\vspace{0.25em}

\noindent\textbf{Implicit Representation for 3D Scene:} Implicit scene representations, notably through Signed Distance Function (SDF), have become prominent in computer graphics and vision for efficiently capturing complex geometry. \cite{lorensen1987marching} firstly proposed a shape reconstruction algorithm from SDF by constructing triangles among 8 voxels. \cite{kazhdan2006poisson} utilizes Octree for accelerated solving of the Poisson equations to compute the indicator function.  Recent studies have explored the implicit representation of continuous 3D shapes as level sets through deep networks. \cite{park2019deepsdf, jiang2020local} learns continuous SDF from point clouds. \cite{mescheder2019occupancy, genova2020local} maps xyz coordinates to occupancy fields, which describes whether a point in 3D space is occupied by a surface or object. \cite{mildenhall2021nerf} represents scenes as neural radiance fields, which are widely used for novel view synthesis \cite{barron2021mip, pumarola2021d}, shape reconstruction \cite{oechsle2021unisurf, wang2021neus} and localization \cite{rosinol2022nerf, deng2023nerf}. 

\vspace{0.25em}

\noindent\textbf{Shape Reconstruction: } Shape reconstruction tasks typically unfold in two distinct scenarios: reconstructing shapes from point clouds or reconstructing shapes from image sequences. For point cloud-based shape reconstruction, \cite{bernardini1999ball} directly connects points to form triangles following a simple ball pivoting rule, without an continuous implicit representation. \cite{carr2003smooth} fits a radial basis function to represent the surface. \cite{kazhdan2006poisson, hou2022iterative} compute the indicator function by iteratively solving the Poisson equations. These methods try to compute an implicit representation, so that the surface can be further extracted through March Cube algorithms \cite{lorensen1987marching}. A fine surface normal estimation is necessary for high reconstruction quality. \cite{lin2022surface} views surface normals as unknown parameters in the Gauss formula to reconstruction shape without normals. \cite{xu2023globally} propose a novel method for globally consistent normal estimation by regularizing the Winding-Number field. Image sequence-based approaches recover 3D shapes from 2D/3D frames instead of a whole point cloud, therefore accurate camera pose estimation is usually important. \cite{niessner2013real, han2018flashfusion, dong2018psdf} perform camera pose estimation and incrementally fuse input frames into distance fields. \cite{ji2017surfacenet, mildenhall2021nerf} take camera parameters and images as input and generate a consistent implicit representation for scene rendering and reconstruction. 

\section{Method}
\begin{figure}[h]
	\centering
	\includegraphics[width=0.85\textwidth, angle=0]{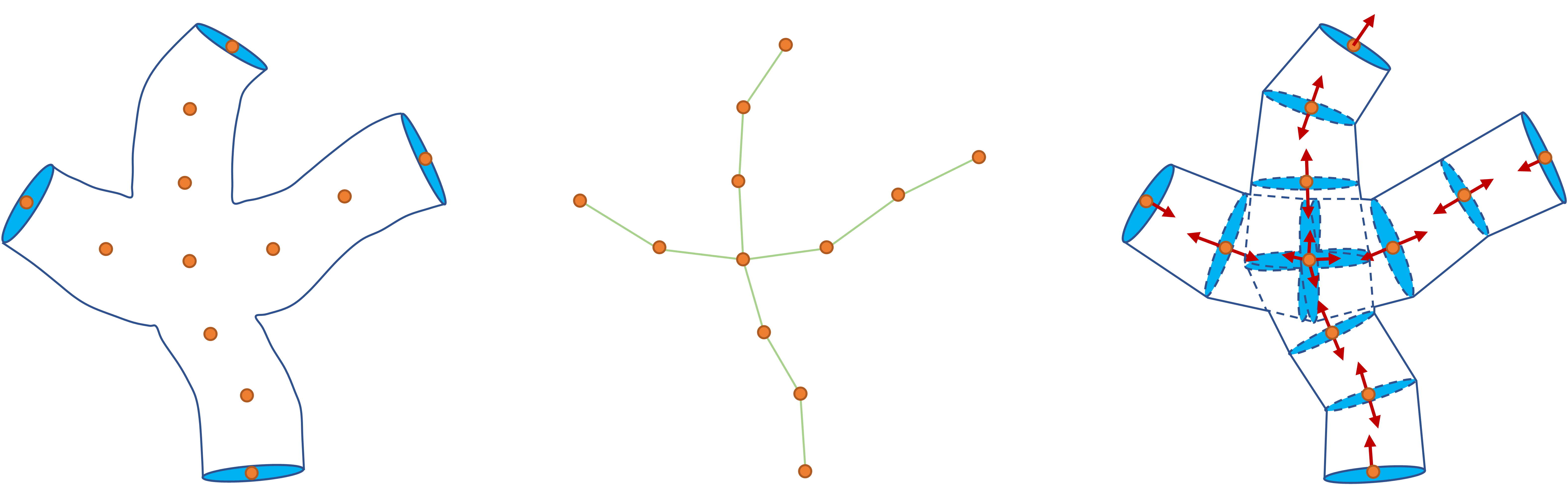}
	\caption{Overview of our shape reconstruction algorithm. Left: original shape and sampled skeletal points. Middle: Constructed graph. Right: recovered shape from slices.}
	\label{fig:unordered}
\end{figure}

Our method tries to recover tubular shape from its skeletal representation that consists of a skeletal point set $\mathcal C = \{\mathbf c_1, \mathbf c_2, ..., \mathbf c_m\}$ and the corresponding radius $\mathbf{r} \in \mathbb{R}^m$. Fig.~\ref{fig:unordered} gives an overview of the reconstruction processing. Firstly, we need to know the connections among the input skeletal points, which is achieved through an adaptive graph construction algorithm as discussed in Sec.~\ref{sec:graph_recon}. Sec.~\ref{sec:sdf_comp} elaborates our geometric algorithm that enables fast computation of signed distance function (SDF). Finally, we introduce how to efficiently represent the shape through voxel hashing strategy and extract the dense model using march cubes algorithm in Sec.~\ref{sec:mesh_extra}. 

\subsection{Graph Construction}
\label{sec:graph_recon}
\begin{algorithm}
\caption{Adaptive Graph Construction}
\label{alg:graph_construction}
\begin{algorithmic}
\Require skeletal points $\mathcal{C}$, number of KNN searched neighbors $k$
\Require distance multiplier $m$  and angle threshold $a_t$
\Ensure constructed graph $\mathcal{G}$
\State $\mathcal{G}\gets$ \textbf{new} Graph
\Comment{Graph initialization}
\State $\mathcal{G}.\mathcal{V} \gets \mathcal{C}$
\State $T\gets$ \textbf{new} KDTree($\mathcal{G}.\mathcal{V}$)
\For{$\mathbf v$ in $\mathcal{G}.\mathcal{V}$ }
\State $\mathcal{I} \gets T$.\text{knn\_search}($\mathbf v$, $k$)
\State $d_\text{min} \gets$  \text{dist}$(\mathbf v, \mathcal{G}.\mathcal{V}[\mathcal{I}[1]])$
\Comment{Calculate distance minimum, note that $\mathcal{I}[0]$ is the index of $\mathbf v$.}
\State $\mathcal{D}_\text{edge} \gets \emptyset$
\Comment{Direction set of edges.}
\For{$i$ in $\mathcal{I}$}
\State $v_i \gets \mathcal{G}.\mathcal{V}[i]$
\If{\text{dist}$(\mathbf v, \mathbf v_i) > m \times d_\text{min}$}
\State \textbf{break}
\EndIf
\State $f \gets$ \textbf{true}
\Comment{Flag of new direction}
\For{$\mathbf d$ in $\mathcal{D}_\text{edge}$}
\If{$\text{angle}(\mathbf v_i - \mathbf v, \mathbf d) <= a_t$}
\State $f \gets$ \textbf{false}
\State \textbf{break}
\EndIf
\EndFor
\Comment{If <$\mathbf v, \mathbf v_i$> a new direction, add it into the edge set}
\If{$f$ is \textbf{true}}
\State $\mathcal{G}.\mathcal{E}.\text{add}(<\mathbf v, \mathbf v_i>)$
\State $\mathcal{D}_{\text{edge}}.\text{add}(\mathbf v_i -\mathbf v)$
\EndIf
\EndFor
\EndFor
\State \textbf{return} $\mathcal{G}$
\end{algorithmic}
\end{algorithm}
To reconstruct the shape, we need to obtain the connections among these skeletal points, which are inherently implied for the ordered skeletal points. 
However, ordered points are only suitable to represent a single curve without any loop or bifurcation. To recover the shape of a complex structure, we need to split the whole structure into multiple segments without loops or bifurcations and reconstruct all the segments separately, which is a tedious work. In this section, we propose an adaptive graph construction algorithm to find connections among unordered skeletal points based on their distance. 

We construct an undirected graph $\mathcal{G} = <\mathcal{V}, \mathcal{E}>$ whose vertices are the input skeletal points: $\mathcal{V} = \mathcal{C}$. Specifically, for each skeletal point $\mathbf{c} \in \mathcal{C}$ we find its $k$ nearest neighbors $\mathbf{c}_1, \mathbf{c}_2,..., \mathbf{c}_k$ that are sorted in ascending order based on their distances to point $\mathbf{c}$. We use $m$ and $a_t$ to denote the distance multiplier and angle threshold, respectively. The distance between $\mathbf c$ and $\mathbf c_i$ is denoted as $d_{\mathbf{c}, \mathbf{c}_i}$. We establish an edge $<\mathbf{c}, \mathbf{c}_i>$, under the condition that $d_{\mathbf{c}, \mathbf{c}_i}$ does not exceed $m$ times the distance between $\mathbf{c}$ and it's nearest neighbor $d_{\mathbf{c}, \mathbf{c}_1}$, and the angles between  $<\mathbf{c}, \mathbf{c}_i>$ and the established edges are greater than $a_t$. This simple strategy obviates the necessity of selecting a suitable distance threshold for each input point cloud. It also avoids the constraint of k-nearest neighbors algorithm-based graph construction that each point need to be connected with a fixed number of points. Alg.~\ref{alg:graph_construction} gives a detailed explanation of the proposed method.


\begin{figure}[h]
	\centering
	\includegraphics[width=0.8\textwidth, angle=0]{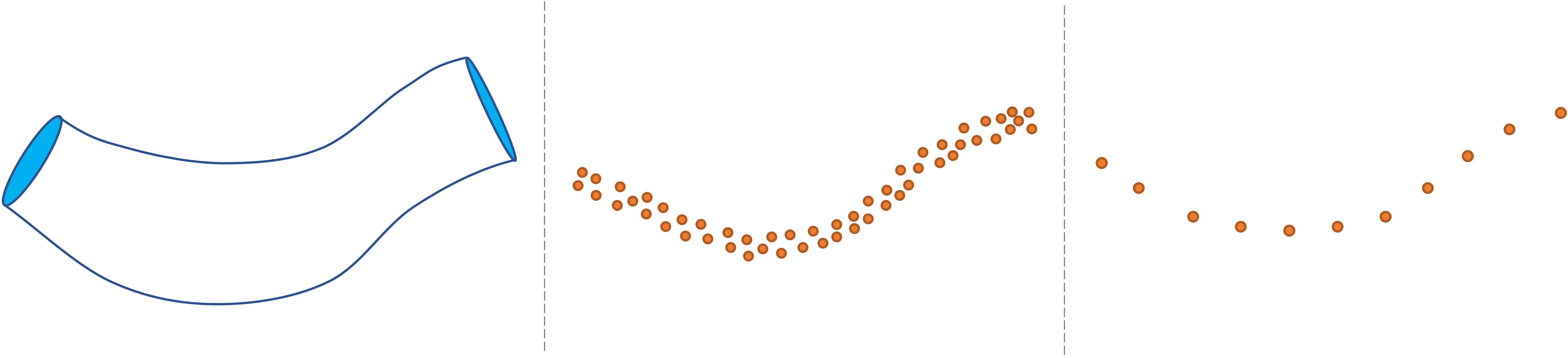}
	\caption{Illustration of different skeletal representations. Left: original shape. Middle: the input skeletal points are a thinner representation of the original shape, but not a string-like structure. Right: the ideal skeletal points forms a string-like structure.}
	\label{fig:string}
\end{figure}

\textbf{Radius-based clustering}: In practice, the skeletal points we acquired may just exhibit a thinned representation of the original shape, and not necessarily be a string-like structure as shown in Fig.~\ref{fig:string}. This may lead to the construction of an extremely complex graph with a large number of redundant edges. To tackle this concern, we employ a pre-processing step that applies a radius-based clustering for the input skeletal points as summarized in Alg.~\ref{alg:radius_clustering}. In specific, we randomly select a skeletal point $\mathbf{c} \in \mathcal{C}$ as a cluster center with corresponding radius $r$. The skeletal points that fall into the ball with center $\mathbf{c}$ and radius $s \cdot r$ are considered to come from the same cluster, which is implemented through radius searching based on KDTree \cite{friedman1977algorithm} and $s$ is the searching strength. We continuously select cluster centers from the un-clustered points and repeat the process until all the skeletal points are successfully clustered. For each cluster, we compute the mean position and radius as a new skeletal point. Through these steps, we successfully reduce the time complexity without a significant decline in reconstruction quality.

\begin{algorithm}
\caption{Radius-based Clustering}
\label{alg:radius_clustering}
\begin{algorithmic}
\Require skeletal points $\mathcal{C}$, radius $\mathbf{r}$ and strength $s$
\Ensure clustered skeletal points $\mathcal{C}'$ and radius $\mathbf{r}'$
\State $\mathcal{C}, \mathbf{r} \gets \text{shuffle}(\mathcal{C}, \mathbf{r})$.
\State $f \gets \text{list}(\vert \mathcal{C}\vert, \textbf{false})$
\Comment{List of flags indicating whether elements are clustered}
\State $T\gets$ \textbf{new} KDTree($\mathcal{C}$)
\State $C', r' \gets \emptyset$

\For{$i$ in \text{range}$(0, \vert\mathcal{C}\vert$)}
\If{$f[i]$ is \textbf{true}}
\State \textbf{continue}
\EndIf
\State $\mathcal{I} \gets T$.\text{radius\_search}($\mathcal{C}[i]$, $s\cdot \mathbf{r}[i]$)
\Comment{Indices of searched neighbors}
\State $\mathbf{c} \gets \mathbf{0}, r \gets 0, n \gets 0$
\For{$j$ in $\mathcal{I}$}
\If{$f[j]$ is \textbf{true}}
\State \textbf{continue}
\EndIf
\State $f[j] \gets$ \textbf{true}
\State $\mathbf{c} \gets \mathbf{c} + \mathcal{C}[j]$
\State $r \gets r + \mathbf{r}[j]$
\State $n \gets n + 1$
\EndFor
\State $\mathcal{C}'$.\text{add}($\mathbf{c} / n $)
\State $\mathbf{r}'$.\text{add}($r / n$)
\Comment{New skeletal point is the center of a cluster}
\EndFor
\State \textbf{return} $\mathcal{C}', \mathbf{r}'$
\end{algorithmic}
\end{algorithm}

\subsection{SDF Computation}
\label{sec:sdf_comp}
\begin{figure}[h]
	\centering
	\includegraphics[width=0.95\textwidth, angle=0]{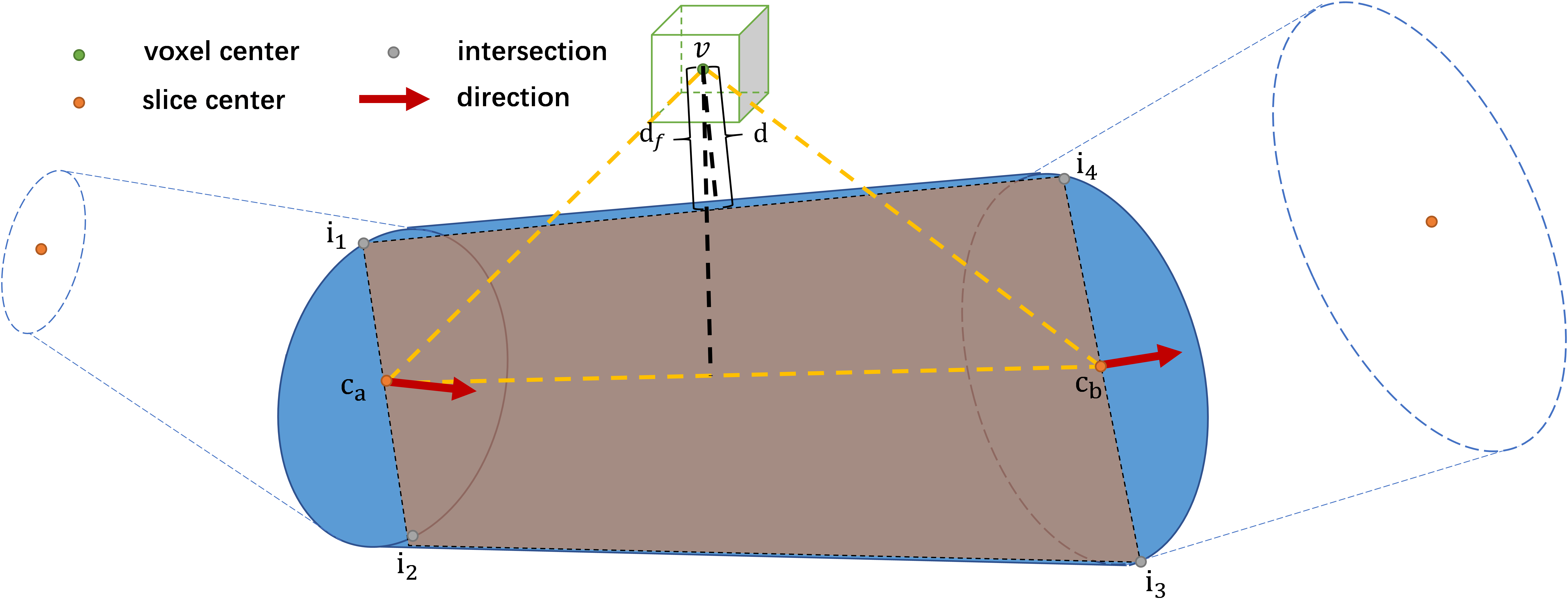}
	\caption{The geometric view of SDF computing.}
	\label{fig:tsdf}
\end{figure}

TSDF represents the geometry of a scene or object in a implicit way. For any point in the space, which is usually the center of a voxel, its signed distance function (SDF) value is defined as the signed distances to the nearest surface. If the point is located inside the object, the sign of distance is negative. TSDF is the SDF truncated by a distance threshold, so that we can only focus on voxels that are near to the surface. Because we use linear interpolation of slices to approximate the original shape, the SDF can be computed through a pure geometric way. Fig.~\ref{fig:tsdf} gives an illustration of SDF computation for a shape between two slices. 

Given two adjacent slices $S_a = (\mathbf{c}_a, \mathbf{n}_a, r_a)$, $S_b = (\mathbf{c}_b, \mathbf{n}_b, r_b)$ and a voxel center point $\mathbf{v}$, we use $f_\text{SDF}(S_a, S_b, \mathbf{v})$ to denote the SDF value of voxel $\mathbf{v}$ for the shape between slices $S_a$ and $S_b$, where $\vert a - b\vert = 1$. The plane $P$ that is uniquely determined by $\mathbf{c}_a$, $\mathbf{c}_b$ and $\mathbf{v}$ intersects with $S_a$ and $S_b$ at four points, denoted as $\mathbf{i}_1$, $\mathbf{i}_2$, $\mathbf{i}_3$ and $\mathbf{i}_4$, respectively. The absolute value of $f_\text{SDF}(S_a, S_b, \mathbf{v})$ is the distance between $\mathbf{v}$ and the nearest surface, which is also the minimum distance between $\mathbf{v}$ and both line segments $\mathbf{i}_1\mathbf{i}_4$ and $\mathbf{i}_2\mathbf{i}_3$. The sign of $f_\text{SDF}(S_a, S_b, \mathbf{v})$ depends on whether the point $\mathbf{v}$ falls into the quadrilateral $\mathbf{i}_1\mathbf{i}_2\mathbf{i}_3\mathbf{i}_4$. 

First, we compute the intersection points of slice $S_a$ and plane $P$. To simplify the computation, we transform the global to local coordinates where the origin is $\mathbf{c}_1$ and the direction vector describing z-axis is $\mathbf{n}_a$. We use $\mathbf{z} = (0, 0, 1)$ to denote the local coordinate of z-direction. The rotation matrix $\mathbf{R}$ from $\mathbf{z}$ to $\mathbf{n}_a$ can be computed by solving a linear equation: 
\begin{equation}
    \mathbf{R}\cdot \mathbf{z} = \mathbf{n}_a, \text{ subject to } \mathbf{R} \cdot \mathbf{R}^{\top} = \mathbf{I},
\end{equation}
where $\mathbf{I}$ is the identity matrix. Here we use a more efficient solution provided by \cite{rodrigues1840lois,rasala1981rodrigues}:
\begin{equation}
    \mathbf{R} = \mathbf{I} + \sin{\theta} \mathbf{K} + (1 - \cos\theta) \mathbf{K}^2.
\end{equation}
In the above, $\mathbf{K} = [\mathbf{z}\times \mathbf{n}_a]_\times$ is a skew-symmetric matrix, $\theta$ is the rotation angle, while $\sin\theta = \Vert \mathbf{z}\times \mathbf{n}_a\Vert$ and $\cos\theta = \mathbf{z} \cdot\mathbf{n}_a$. The translation from origin $(0,0,0)$ to $\mathbf{c}_a$ is $\mathbf{c}_a$, therefore we can easily get the transformation matrix from local to global coordinates $\mathbf{T} = \left[\mathbf{R} \vert \mathbf{c}_a\right]$ whose inverse is the transformation matrix from global to local coordinates.

The normal vector $\mathbf{n}_P$ of plane $P$ can be computed through following equation:
\begin{equation}
    \mathbf{n}_P = \frac{(\mathbf{c}_a - \mathbf{v}) \times (\mathbf{c}_b - \mathbf{v}) }{\Vert (\mathbf{c}_a - \mathbf{v}) \times (\mathbf{c}_b - \mathbf{v}) \Vert}.
\end{equation}
The intersections $\mathbf{i}_1$ and $\mathbf{i}_2$ of plane $P$ and slice $S_a$ can be calculated by solving following system of equations:
\begin{align}
        &\mathbf{n}_P^l \cdot \mathbf{p} = 0\label{eq:plane}\\
        &\Vert \mathbf{p} \Vert = r_a\label{eq:circle}\\
        &\mathbf{p}_z = 0\label{eq:xy}.
\end{align}
Equ.~\ref{eq:plane} represents the plane $P$, where $\mathbf{n}_P^l = \mathbf{R}^{-1}\cdot \mathbf{n}_P$. Equ.~\ref{eq:circle} and Equ.~\ref{eq:xy} represent the slice $S_a$. The transformation from global to local coordinates actually degrades the solutions from to 3D into 2D, because all points have a z-axis value of zero. The solutions of above system are denoted as $\mathbf{i}_1^l$ and $\mathbf{i}_2^l$, respectively, and we have:
\begin{equation}
    \mathbf{i}_1^l = - \mathbf{i}_2^l.
\end{equation}
Transform the solutions back to global coordinates, we have:
\begin{equation}
    \mathbf{i}_1 = \mathbf T\cdot \mathbf{i}_1^l, \mathbf{i}_2 = \mathbf T\cdot \mathbf{i}_2^l.
\end{equation}
The point that is on the same side of line $\mathbf{c}_a\mathbf{c}_b$ with $\mathbf{v}$ is denoted as $\mathbf{i}_a$, which is an end point of the line segment closer to $\mathbf{v}$. 

Vice versa, we compute the $\mathbf{i}_b$ for slice $S_b$. The value of SDF can be calculated as follows:
\begin{equation}
    f_\text{SDF}(S_a, S_b, \mathbf{v}) =\left\{\begin{matrix}
        -\text{dist}(\mathbf{i_a}\mathbf{i_b}, \mathbf{v}), & \text{if } \mathbf{v} \text{ is in  the quadrilateral } \mathbf{c}_a \mathbf{c}_b \mathbf{i}_b \mathbf{i}_a;\\
        \text{dist}(\mathbf{i_a}\mathbf{i_b}, \mathbf{v}), & \text{otherwise}.
    \end{matrix}\right.
\end{equation}
where $\text{dist}(\cdot, \cdot)$ returns the distance between a line segment and and a point.

The function $f_\text{SDF}(S_a, S_b, \mathbf{v})$ calculates the SDF of voxel $ \mathbf{v}$ for partial shape between two adjacent slices $S_a$ and $S_b$. Given graph $\mathcal{G}$, the SDF of $\mathbf{v}$ for the whole structure is defined as:
\begin{equation}
    f_\text{SDF}(\mathbf{v}) = \min\{ \min_{e \in \mathcal{G.E}} f_\text{SDF}(S_{e[0]}, S_{e[1]}, \mathbf{v}), d_{\text{ball}}\},
\end{equation}
where $e[0]$ and $e[1]$ is the indices of two endpoints of edge $e$, $d_{\text{ball}}$ is a smoothing factor, which is defined as the signed distance to the ball determined by the nearest skeletal point $\mathbf{c}_n$: $d_{\text{ball}}= \Vert \mathbf{v} - \mathbf{c}_n \Vert - r_n$.

\begin{wrapfigure}{r}{0.45\textwidth} 
  \centering
  \vspace{-1.5em} 
  \includegraphics[width=\linewidth]{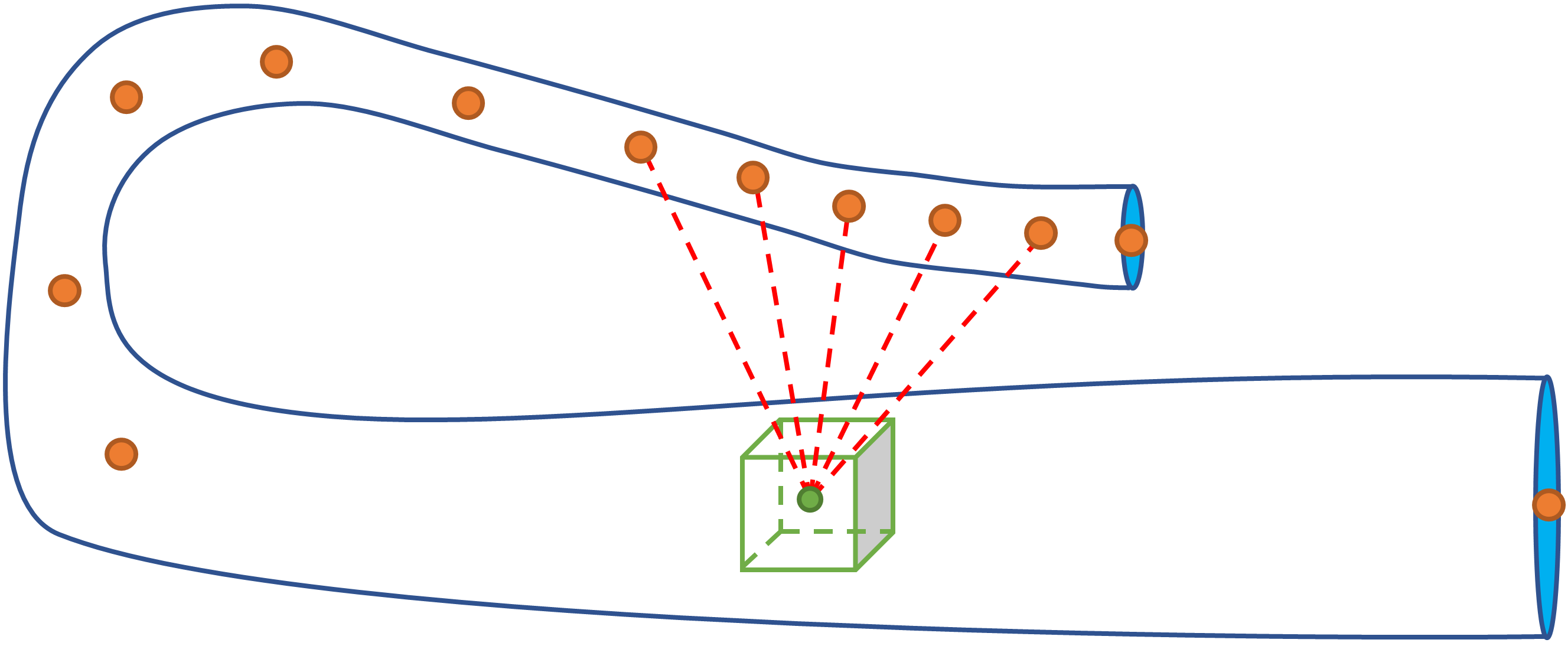} 
  \vspace{-1em} 
  \caption{A failure case when faced with poor sampling.} 
  \vspace{-1em} 
  \label{fig:failure}
\end{wrapfigure}

To obtain the SDF of $\mathbf{v}$ for the whole shape, we need to compute SDF values of $\mathbf{v}$ for all edges in the graph, which is time-consuming and inefficient. In our implementation, we only compute the SDF values for 5 nearest edges. This is based on a simple assumption: the skeletal points is well sampled so that far skeletal points have little effect on the SDF computing. Fig.~\ref{fig:failure} shows a failure case if the assumption is not satisfied, in which all 5 nearest skeletal points of a voxel inside the tube are outside the corresponding partial shape, leading to an incorrect SDF computation. 

\textbf{Fast SDF Approximation:} The aforementioned method computes the distance between a voxel center and it's nearest surface that is denoted as $d$ in Fig.~\ref{fig:tsdf}. A more fast way to approximate the SDF value for a shape between two adjacent slices $(\mathbf{c}_a, \mathbf{n}_a, r_a)$ and $(\mathbf{c}_b, \mathbf{n}_b, r_b)$ is to assume $\mathbf{n}_a = \mathbf{n}_b = \mathbf{c}_b - \mathbf{c}_a$ and compute $d_\text{f}$. For a given voxel center $v$, we project it to line segment $\mathbf{c}_a \mathbf{c}_b$ at a new skeletal point $\mathbf{c}_v$, whose radius can be calculated by a simple interpolation:
$$
r_v = \frac{r_b \cdot \Vert\mathbf{c}_v - \mathbf{c}_a \Vert   + r_a \cdot \Vert \mathbf{c}_b - \mathbf{c}_v \Vert}{\Vert \mathbf{c}_b - \mathbf{c}_a \Vert}.
$$
And then we have:
$$
d_{\text{f}} = \Vert \mathbf{v} - \mathbf{c}_v \Vert - r_v.
$$

\begin{wrapfigure}{l}{0.45\textwidth} 
  \centering
  \vspace{-1.5em} 
  \includegraphics[width=\linewidth]{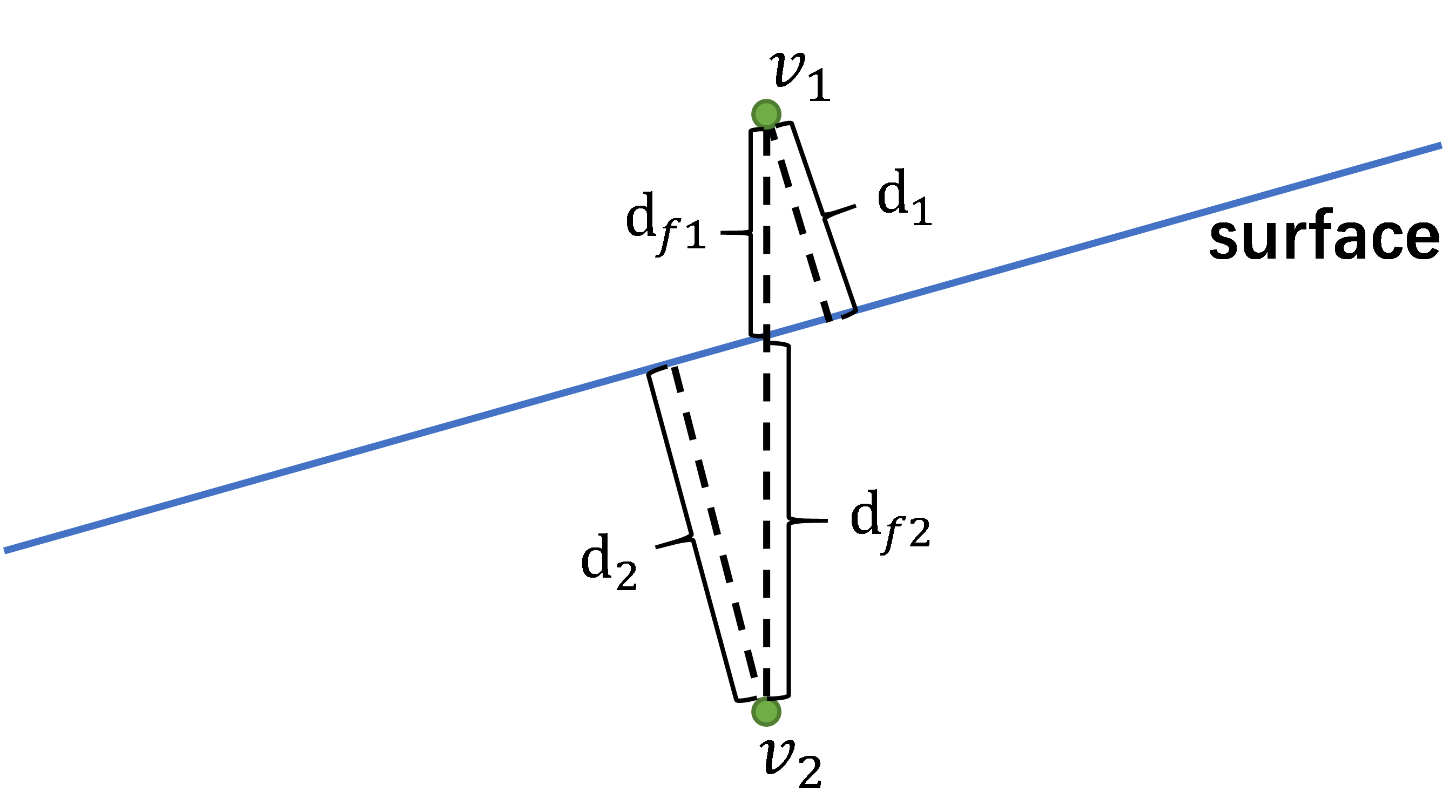} 
  \vspace{-1em} 
  \caption{Surface extraction using different SDF values.} 
  \vspace{-1.5em} 
  \label{fig:fast}
\end{wrapfigure}

Note that the approximated SDF value calculated through fast computation still generates an accurate surface under the equal normal assumption. As shown in Fig.~\ref{fig:fast}, $d_1$ and $d_2$ represent the SDF values of two adjacent voxel centers $\mathbf{v}_1$ and $\mathbf{v}_2$, respectively, while $d_{f1}$ and $d_{f2}$ represent the approximated SDF values. According to the principle of similar triangles, we have $\frac{d_1}{d_2} = \frac{d_{f1}}{d_{f2}}$. Because the surface point is the weighted interpolation of voxel centers which can be formulated as:
$$
\mathbf{v} = \frac{d_2}{d_1 + d_2} \cdot \mathbf{v}_1 + \frac{d_1}{d_1 + d_2} \cdot \mathbf{v}_2 = \frac{1}{\frac{d_1}{d_2} + 1} \cdot \mathbf{v}_1 + \frac{1}{1 + \frac{d_2}{d_1}} \cdot \mathbf{v}_2,
$$
using the approximated SDF values generates the same result as using true SDF values. Note that, providing an algorithm to compute real SDF values is not without merit, as real SDF values are more useful in some tasks \cite{park2019deepsdf}.
\subsection{Voxel Hashing and Mesh Extraction}
\label{sec:mesh_extra}
Instead of computing the SDF value for each voxel in the bounding box, we use a spatial hashing table \cite{niessner2013real,han2018flashfusion} to store sparse cubes that near to the object, which greatly reduces the required memory and computational complexity. The space is represented as cubes, while each cube consists of $8 \times 8 \times 8$ voxels. We compute a bounding box for the input skeletal points, any cube that is located within the bounding box or has an overlapping region with the bounding box is referred to a candidate cube. For each candidate cube, we calculate the SDF values for its 8 corner voxels. If the minimum absolute value among these values is smaller than the truncated distance, or if the signs of these values are not the same, we consider that there might be some parts of surface in the cube, and the cube is added into the hash table. The truncated SDF is then calculated for all voxels in selected cubes. Finally, we use the multi-threaded marching cubes algorithm \cite{lorensen1987marching} to extract triangle meshes. The corresponding mask volume can also be easily obtained from the TSDF by identifying whether the voxels are located inside the object. Overall, the proposed shape reconstruction algorithm is summarized in Arg.~\ref{alg:unorder}. The default values of mentioned parameters during the processing are listed in Tab.~\ref{tab:para}.

\begin{algorithm}[h]
\caption{Shape reconstruction from unordered skeletal points}\label{alg:unorder}
\begin{algorithmic}
\Require skeletal points $\mathcal{C}$, radius $\mathbf{r}$, number of KNN searched neighbors $k$
\Require clustering strength $s$, voxel resolution $l$ and truncated distance $d_t$
\Require distance multiplier $m$, angle threshold $a_t$
\Ensure $M$: reconstructed triangle mesh
\State $\mathcal{C}, \mathbf{r} \gets \text{radius\_clustering}(\mathcal{C}, \mathbf{r}, s)$ 
\State $\mathcal{G} \gets \text{adaptive\_graph\_construction}(\mathcal{C}, k, m, a_t)$
\Function \text{compute\_sdf}$(\mathbf v)$: 
\Comment{SDF computation}
\State $\mathcal{I} \gets T$.\text{knn\_search}($\mathbf v$, $k$)
\State $f_\text{min} \gets d_t$
\For{$i$ in $\mathcal{I}$}
\State $\mathbf v_i \gets \mathcal{G}.\mathcal{V}[i]$ 
\For{$\mathbf v_j$ in $\mathcal{G}.\mathcal{E}(\mathbf v_i)$}
\State $f \gets\text{compute\_sdf}(\mathbf v, \mathbf v_i, \mathbf v_j)$
\Comment{As mentioned in Sec.~\ref{sec:sdf_comp}}
\If{$f < f_\text{min}$}
\State $f_\text{min} \gets f$
\EndIf
\EndFor
\EndFor
\State $d_{c} \gets \text{dist}\left(\mathbf v, \mathcal{G}.\mathcal{V}\left[\mathcal{I}[0]\right]\right) - \mathbf{r}\left[\mathcal{I}[0]\right]$
\Comment{Distance between $\mathbf{v}$ and the surface of the closest ball}
\If{$f_\text{min} > d_c$}
\State $f_\text{min} \gets d_c$
\EndIf
\State \textbf{return} $f_\text{min}$
\EndFunction
\State $\mathcal{H}\gets\text{spatial\_hashing}(\mathcal{C},\text{compute\_sdf})$
\Comment{Store selected cubes in a spatial hashing table}
\For{$v$ in $\mathcal{H}$}
\State $H[\mathbf v]$.sdf $\gets$ \text{compute\_sdf}($\mathbf v$)
\EndFor
\State $M \gets$\text{marching\_cubes}($\mathcal{H}$) 
\Comment{Mesh extraction through Marching Cubes algorithm}
\State \textbf{return} $M$
\end{algorithmic}
\end{algorithm}

\begin{table*}[h]
\centering
\caption{Parameter table. Default values are set empirically.}
\label{tab:para}
\begin{tabular}{c|l|c|c}
\hline
Algorithm & Parameter                   & Notation  & Default Value\\ \hline
Radius clustering& Clustering strength & $s$ & 0.75 \\    \hline
\multirow{3}{*}{Graph reconstruction} & Number of KNN searched neighbors & $k$ & 5 \\    
& Distance multiplier & $m$ & 2.5 \\   
& Angle threshold & $a_t$ & 75 \\    \hline
\multirow{3}{*}{SDF computation} & Number of KNN searched neighbors & $k$ & 5\\
& Voxel resolution & $l$ & 0.025 \\    
& Truncated distance & $d_t$ & 0.1 \\ \hline
\end{tabular}
\vspace{-1em}
\end{table*}

\section{Experimental Results}
\subsection{Datasets}

\begin{figure}[h]
	\centering
	\includegraphics[width=0.95\textwidth, angle=0]{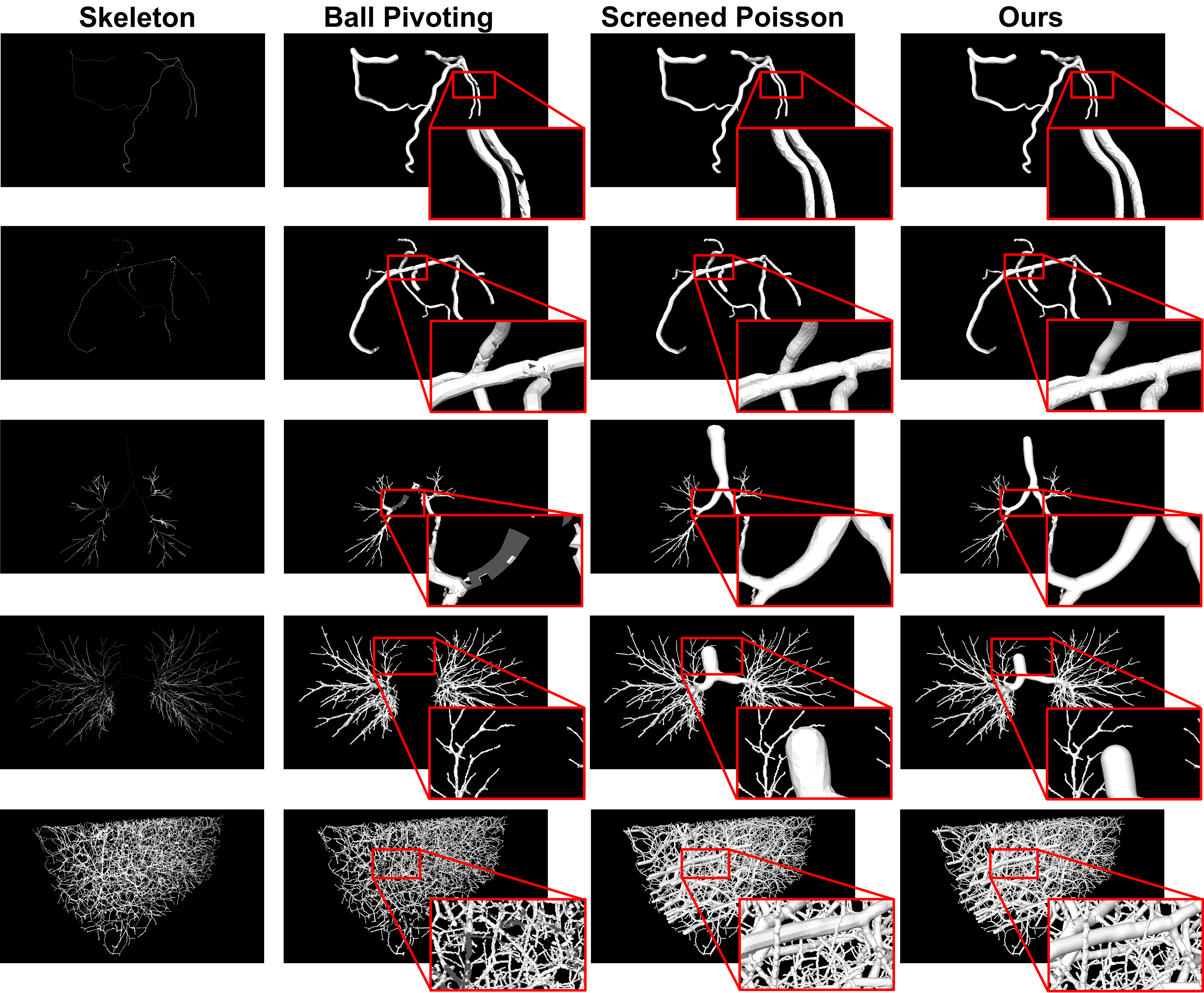}
	\caption{Qualitative experimental results. From top to bottom each row shows reconstruction results for datasets CAT08, ImageCAS, ATM22, PARSE2022 and DeepVesselNet, respectively. Columns from left to right are visualizations of skeletal points and reconstructed shapes using different methods. The recovered shapes by our method are more complete and accurate, achieving the best reconstruction quality.}
	\label{fig:res}
\end{figure}

We evaluate our method on five public datasets. CAT08 \cite{schaap2009standardized} is a coronary artery center-line extraction dataset which contains 8 CCTA volumes with carefully labeled center-lines. ImagesCAS \cite{zeng2022imagecas} is a large coronary artery segmentation dataset that contains 1000 3D images with sizes of 512 $\times$ 512 $\times$ (206-275). ATM22 \cite{zhang2023multi} is a pulmonary airway segmentation dataset and contains 500 CT scans with sizes of 512 $\times$ 512 $\times$ (84-879). PARSE2022 \cite{maurya2022parse} contains 200 3D volumes with sizes of 512 $\times$ 512 $\times$ (228-376)  for  pulmonary artery segmentation. DeepVesselNet \cite{ tetteh2020deepvesselnet} contains 100 synthetic 3D images for complicated vessel segmentation. The images have a size of 325 $\times$ 304 $\times$ 600 voxels. All these segmentation datasets only contains labeled masks, and their skeletons are calculated through a binary thinning algorithm \cite{zhang1984fast}. 
\begin{table*}[h]
\centering
\caption{Quantitative experimental results.}
\label{tab:acc}
\resizebox{0.95\columnwidth}{!}{
\begin{tabular}{cl|ccc|c}
\hline
Dataset                     & Method  & Dice Score & Radius Difference & Center Agreement & Average Running Time (s) \\ \hline
\multirow{4}{*}{CAT08 (1496)} & Ball Pivoting & - & 0.2286 & 0.9194 & \textbf{0.18} \\    
                            & Poisson  & - & 0.1887 & 0.9999 & 0.61 \\ 
                            & Ours & - & 0.1013 & 1.0000 & 0.54\\ 
                            & Ours (fast)  & - & \textbf{0.1010} & \textbf{1.0000} & 0.24\\ \hline
\multirow{4}{*}{ImageCAS (1507)} & Ball Pivoting & 0.7007  & 0.7619 & 0.6999 & 0.22 \\   
                            & Poisson & 0.7548  & \textbf{0.4614} & 0.9939 & 0.76\\ 
                            & Ours & 0.7566 & 0.5983 & 0.9994 & 0.39\\ 
                            & Ours (fast) & \textbf{0.7638}  & 0.5511 & \textbf{0.9997} & \textbf{0.22}\\ \hline
\multirow{4}{*}{ATM22 (2688)}   & Ball Pivoting & 0.3619  & 1.2396 & 0.7040 & 2.11 \\   
                            & Poisson  & 0.6805 & 0.7867 & 0.9929 & 1.89\\ 
                            & Ours & 0.6882 & 0.4884 & 0.9992 & 1.56\\ 
                            & Ours (fast)  & \textbf{0.7045}  & \textbf{0.4048} & \textbf{0.9997} & \textbf{0.93}\\ \hline
\multirow{4}{*}{PARSE2022 (8885)}   & Ball Pivoting & 0.2810 & 0.8912 & 0.7527 & 8.50 \\    
                            & Poisson & 0.6263 & 0.6415 &  0.9915 & 4.87\\ 
                            & Ours & 0.6347 & 0.4883 & 0.9985 & 4.63\\ 
                            & Ours (fast)  & \textbf{0.6516} & \textbf{0.3773} & \textbf{0.9994} & \textbf{3.21}\\ \hline
\multirow{4}{*}{DeepVesselNet (62519)}   & Ball Pivoting &0.2960& 0.6602 & 0.7943 & 73.32 \\   
                            & Poisson & 0.7387  & 0.3239 & 0.9987 & 32.13\\ 
                            & Ours & 0.7473 & 0.3394 & 0.9998 & 25.96\\ 
                            & Ours (fast) & \textbf{0.7498} & \textbf{0.2819} & \textbf{0.9999} & \textbf{19.54}\\ \hline
\end{tabular}}
\vspace{-1em}
\end{table*}
\subsection{Evaluation Metrics}
We use three metrics to measure the reconstruction accuracy.  We recover mask volume from reconstructed shape and compute \textbf{Dice score} between ground-truth and recovered mask for segmentation datasets. \textbf{Radius difference} is defined as the average different between the re-computed radius of recovered shape and original radius. \textbf{Center agreement} calculates the percentage of skeletal points that are located inside the recovered shape. Additionally, we compare the efficiency of different reconstruction algorithms through the \textbf{average running time}.

\subsection{Qualitative and Quantitative Analysis}
We compare our algorithm with two previous works. Ball Pivoting \cite{bernardini1999ball} tries to connect the surface points to form triangles, while Screened Poisson \cite{kazhdan2013screened} calculates the indicator function by solving the regularized Poisson equation. Both of them are shape reconstruction algorithms from surface points and cannot be directly applied on skeletal representations. Therefore, for each slice we generate 10 surface points to run the baseline algorithms. Note that the slices are generated from constructed graph as mentioned in Sec.~\ref{sec:graph_recon}.

Tab.~\ref{tab:acc} shows the quantitative results, where we introduce the accuracy and performance of our reconstruction algorithms using real and approximated sdf values, denoted as \textbf{ours} and \textbf{ours (fast)}, respectively . Our methods demonstrate high superiority in reconstruction accuracy. While the mask and shape recovered by ours (fast) maintains the highest consistency with original mask and skeleton, the reconstruction accuracy with real sdf values slightly decreases, possibly due to the precision loss incurred during the computation process. Meanwhile, the proposed methods are generally more efficient. Ours (fast) is 2.38$\times$ faster than Ball Pivoting \cite{bernardini1999ball} and 1.67$\times$ faster than Screened Poisson \cite{kazhdan2013screened}, respectively. It's worth noting that the shapes recovered by our methods also have much smaller volumes than the ones by Screened Poisson \cite{kazhdan2013screened} while maintaining a higher reconstruction quality.

\section{Conclusion}
We propose a novel algorithmic pipeline to reconstruct the tubular shape from skeletal representation. Specifically, we use a radius-based clustering as pre-processing to reduce point redundancy and obtain string-like skeletal points. Based on the simple assumption that tubular shape can be recovered through a linear interpolation of slices, we analyze the geometric solution of SDF computing and propose a fast SDF approximation. Finally, we use a parallel marching cube algorithm to extract surface models. Experiments shows that our method achieves superior reconstruction quality with lower computational complexity compared to other methods.

\textbf{Acknowledgements}: This work is supported in part by the Natural Science Foundation of China (Grant 62371270), the Major Key Project of PCL (Grant PCL2023A09, Pengcheng Laboratory), and Shenzhen Key Laboratory of Ubiquitous Data Enabling (No.ZDSYS20220527171406015).

\bibliographystyle{unsrt}
\bibliography{references}  

\begin{thebibliography}{10}

\bibitem{carr2003smooth}
Jonathan~C Carr, Richard~K Beatson, Bruce~C McCallum, W~Richard Fright, Tim~J
  McLennan, and Tim~J Mitchell.
\newblock Smooth surface reconstruction from noisy range data.
\newblock In {\em Proceedings of the 1st international conference on Computer
  graphics and interactive techniques in Australasia and South East Asia},
  pages 119--ff, 2003.

\bibitem{kazhdan2006poisson}
Michael Kazhdan, Matthew Bolitho, and Hugues Hoppe.
\newblock Poisson surface reconstruction.
\newblock In {\em Proceedings of the fourth Eurographics symposium on Geometry
  processing}, volume~7, page~0, 2006.

\bibitem{hou2022iterative}
Fei Hou, Chiyu Wang, Wencheng Wang, Hong Qin, Chen Qian, and Ying He.
\newblock Iterative poisson surface reconstruction (ipsr) for unoriented
  points.
\newblock {\em arXiv preprint arXiv:2209.09510}, 2022.

\bibitem{niessner2013real}
Matthias Nie{\ss}ner, Michael Zollh{\"o}fer, Shahram Izadi, and Marc
  Stamminger.
\newblock Real-time 3d reconstruction at scale using voxel hashing.
\newblock {\em ACM Transactions on Graphics (ToG)}, 32(6):1--11, 2013.

\bibitem{han2018flashfusion}
Lei Han and Lu~Fang.
\newblock Flashfusion: Real-time globally consistent dense 3d reconstruction
  using cpu computing.
\newblock In {\em Robotics: Science and Systems}, volume~1, page~7, 2018.

\bibitem{mildenhall2021nerf}
Ben Mildenhall, Pratul~P Srinivasan, Matthew Tancik, Jonathan~T Barron, Ravi
  Ramamoorthi, and Ren Ng.
\newblock Nerf: Representing scenes as neural radiance fields for view
  synthesis.
\newblock {\em Communications of the ACM}, 65(1):99--106, 2021.

\bibitem{dong2018psdf}
Wei Dong, Qiuyuan Wang, Xin Wang, and Hongbin Zha.
\newblock Psdf fusion: Probabilistic signed distance function for on-the-fly 3d
  data fusion and scene reconstruction.
\newblock In {\em Proceedings of the European conference on computer vision
  (ECCV)}, pages 701--717, 2018.

\bibitem{huang2020pf}
Zitian Huang, Yikuan Yu, Jiawen Xu, Feng Ni, and Xinyi Le.
\newblock Pf-net: Point fractal network for 3d point cloud completion.
\newblock In {\em Proceedings of the IEEE/CVF conference on computer vision and
  pattern recognition}, pages 7662--7670, 2020.

\bibitem{wen2020point}
Xin Wen, Tianyang Li, Zhizhong Han, and Yu-Shen Liu.
\newblock Point cloud completion by skip-attention network with hierarchical
  folding.
\newblock In {\em Proceedings of the IEEE/CVF Conference on Computer Vision and
  Pattern Recognition}, pages 1939--1948, 2020.

\bibitem{kraevoy2005template}
Vladislav Kraevoy and Alla Sheffer.
\newblock Template-based mesh completion.
\newblock In {\em Symposium on Geometry Processing}, volume 385, pages 13--22,
  2005.

\bibitem{dai2017shape}
Angela Dai, Charles Ruizhongtai~Qi, and Matthias Nie{\ss}ner.
\newblock Shape completion using 3d-encoder-predictor cnns and shape synthesis.
\newblock In {\em Proceedings of the IEEE conference on computer vision and
  pattern recognition}, pages 5868--5877, 2017.

\bibitem{choi1997mathematical}
Hyeong~In Choi, Sung~Woo Choi, and Hwan~Pyo Moon.
\newblock Mathematical theory of medial axis transform.
\newblock {\em pacific journal of mathematics}, 181(1):57--88, 1997.

\bibitem{zwettler2008accelerated}
Gerald Zwettler, Franz Pfeifer, Roland Swoboda, and Werner Backfrieder.
\newblock Accelerated skeletonization algorithm for tubular structures in large
  datasets by randomized erosion.
\newblock In {\em International Conference on Computer Vision Theory and
  Applications}, volume~2, pages 74--81. SCITEPRESS, 2008.

\bibitem{wang2020deep}
Yan Wang, Xu~Wei, Fengze Liu, Jieneng Chen, Yuyin Zhou, Wei Shen, Elliot~K
  Fishman, and Alan~L Yuille.
\newblock Deep distance transform for tubular structure segmentation in ct
  scans.
\newblock In {\em Proceedings of the IEEE/CVF Conference on Computer Vision and
  Pattern Recognition}, pages 3833--3842, 2020.

\bibitem{shit2021cldice}
Suprosanna Shit, Johannes~C Paetzold, Anjany Sekuboyina, Ivan Ezhov, Alexander
  Unger, Andrey Zhylka, Josien~PW Pluim, Ulrich Bauer, and Bjoern~H Menze.
\newblock cldice-a novel topology-preserving loss function for tubular
  structure segmentation.
\newblock In {\em Proceedings of the IEEE/CVF Conference on Computer Vision and
  Pattern Recognition}, pages 16560--16569, 2021.

\bibitem{metz20083d}
Coert Metz, Michiel Schaap, Theo van Walsum, Alina van~der Giessen, Annick
  Weustink, Nico Mollet, Gabriel Krestin, and Wiro Niessen.
\newblock 3d segmentation in the clinic: A grand challenge ii-coronary artery
  tracking.
\newblock {\em Insight Journal}, 1(5):6, 2008.

\bibitem{schaap2009standardized}
Michiel Schaap, Coert~T Metz, Theo van Walsum, Alina~G van~der Giessen,
  Annick~C Weustink, Nico~R Mollet, Christian Bauer, Hrvoje Bogunovi{\'c},
  Carlos Castro, Xiang Deng, et~al.
\newblock Standardized evaluation methodology and reference database for
  evaluating coronary artery centerline extraction algorithms.
\newblock {\em Medical image analysis}, 13(5):701--714, 2009.

\bibitem{xing2001numerical}
HL~Xing and A~Makinouchi.
\newblock Numerical analysis and design for tubular hydroforming.
\newblock {\em International Journal of Mechanical Sciences}, 43(4):1009--1026,
  2001.

\bibitem{hu2019topology}
Xiaoling Hu, Fuxin Li, Dimitris Samaras, and Chao Chen.
\newblock Topology-preserving deep image segmentation.
\newblock {\em Advances in neural information processing systems}, 32, 2019.

\bibitem{luo2021design}
Chen Luo, Chuan~Zhen Han, Xiang~Yu Zhang, Xue~Gang Zhang, Xin Ren, and Yi~Min
  Xie.
\newblock Design, manufacturing and applications of auxetic tubular structures:
  A review.
\newblock {\em Thin-Walled Structures}, 163:107682, 2021.

\bibitem{gharleghi2022towards}
Ramtin Gharleghi, Nanway Chen, Arcot Sowmya, and Susann Beier.
\newblock Towards automated coronary artery segmentation: A systematic review.
\newblock {\em Computer Methods and Programs in Biomedicine}, page 107015,
  2022.

\bibitem{huang2013l1}
Hui Huang, Shihao Wu, Daniel Cohen-Or, Minglun Gong, Hao Zhang, Guiqing Li, and
  Baoquan Chen.
\newblock L1-medial skeleton of point cloud.
\newblock {\em ACM Trans. Graph.}, 32(4):65--1, 2013.

\bibitem{qin2019mass}
Hongxing Qin, Jia Han, Ning Li, Hui Huang, and Baoquan Chen.
\newblock Mass-driven topology-aware curve skeleton extraction from incomplete
  point clouds.
\newblock {\em IEEE transactions on visualization and computer graphics},
  26(9):2805--2817, 2019.

\bibitem{lin2021point2skeleton}
Cheng Lin, Changjian Li, Yuan Liu, Nenglun Chen, Yi-King Choi, and Wenping
  Wang.
\newblock Point2skeleton: Learning skeletal representations from point clouds.
\newblock In {\em Proceedings of the IEEE/CVF conference on computer vision and
  pattern recognition}, pages 4277--4286, 2021.

\bibitem{Zhang_2023_BMVC}
Guoqing Zhang, Caixia Dong, and Yang Li.
\newblock Topology-preserving hard pixel mining for tubular structure
  segmentation.
\newblock In {\em 34th British Machine Vision Conference 2023, {BMVC} 2023,
  Aberdeen, UK, November 20-24, 2023}. BMVA, 2023.

\bibitem{lorensen1987marching}
William~E Lorensen and Harvey~E Cline.
\newblock Marching cubes: A high resolution 3d surface construction algorithm.
\newblock {\em ACM siggraph computer graphics}, 21(4):163--169, 1987.

\bibitem{park2019deepsdf}
Jeong~Joon Park, Peter Florence, Julian Straub, Richard Newcombe, and Steven
  Lovegrove.
\newblock Deepsdf: Learning continuous signed distance functions for shape
  representation.
\newblock In {\em Proceedings of the IEEE/CVF conference on computer vision and
  pattern recognition}, pages 165--174, 2019.

\bibitem{jiang2020local}
Chiyu Jiang, Avneesh Sud, Ameesh Makadia, Jingwei Huang, Matthias Nie{\ss}ner,
  Thomas Funkhouser, et~al.
\newblock Local implicit grid representations for 3d scenes.
\newblock In {\em Proceedings of the IEEE/CVF Conference on Computer Vision and
  Pattern Recognition}, pages 6001--6010, 2020.

\bibitem{mescheder2019occupancy}
Lars Mescheder, Michael Oechsle, Michael Niemeyer, Sebastian Nowozin, and
  Andreas Geiger.
\newblock Occupancy networks: Learning 3d reconstruction in function space.
\newblock In {\em Proceedings of the IEEE/CVF conference on computer vision and
  pattern recognition}, pages 4460--4470, 2019.

\bibitem{genova2020local}
Kyle Genova, Forrester Cole, Avneesh Sud, Aaron Sarna, and Thomas Funkhouser.
\newblock Local deep implicit functions for 3d shape.
\newblock In {\em Proceedings of the IEEE/CVF Conference on Computer Vision and
  Pattern Recognition}, pages 4857--4866, 2020.

\bibitem{barron2021mip}
Jonathan~T Barron, Ben Mildenhall, Matthew Tancik, Peter Hedman, Ricardo
  Martin-Brualla, and Pratul~P Srinivasan.
\newblock Mip-nerf: A multiscale representation for anti-aliasing neural
  radiance fields.
\newblock In {\em Proceedings of the IEEE/CVF International Conference on
  Computer Vision}, pages 5855--5864, 2021.

\bibitem{pumarola2021d}
Albert Pumarola, Enric Corona, Gerard Pons-Moll, and Francesc Moreno-Noguer.
\newblock D-nerf: Neural radiance fields for dynamic scenes.
\newblock In {\em Proceedings of the IEEE/CVF Conference on Computer Vision and
  Pattern Recognition}, pages 10318--10327, 2021.

\bibitem{oechsle2021unisurf}
Michael Oechsle, Songyou Peng, and Andreas Geiger.
\newblock Unisurf: Unifying neural implicit surfaces and radiance fields for
  multi-view reconstruction.
\newblock In {\em Proceedings of the IEEE/CVF International Conference on
  Computer Vision}, pages 5589--5599, 2021.

\bibitem{wang2021neus}
Peng Wang, Lingjie Liu, Yuan Liu, Christian Theobalt, Taku Komura, and Wenping
  Wang.
\newblock Neus: Learning neural implicit surfaces by volume rendering for
  multi-view reconstruction.
\newblock {\em arXiv preprint arXiv:2106.10689}, 2021.

\bibitem{rosinol2022nerf}
Antoni Rosinol, John~J Leonard, and Luca Carlone.
\newblock Nerf-slam: Real-time dense monocular slam with neural radiance
  fields.
\newblock {\em arXiv preprint arXiv:2210.13641}, 2022.

\bibitem{deng2023nerf}
Junyuan Deng, Qi~Wu, Xieyuanli Chen, Songpengcheng Xia, Zhen Sun, Guoqing Liu,
  Wenxian Yu, and Ling Pei.
\newblock Nerf-loam: Neural implicit representation for large-scale incremental
  lidar odometry and mapping.
\newblock In {\em Proceedings of the IEEE/CVF International Conference on
  Computer Vision}, pages 8218--8227, 2023.

\bibitem{bernardini1999ball}
Fausto Bernardini, Joshua Mittleman, Holly Rushmeier, Cl{\'a}udio Silva, and
  Gabriel Taubin.
\newblock The ball-pivoting algorithm for surface reconstruction.
\newblock {\em IEEE transactions on visualization and computer graphics},
  5(4):349--359, 1999.

\bibitem{lin2022surface}
Siyou Lin, Dong Xiao, Zuoqiang Shi, and Bin Wang.
\newblock Surface reconstruction from point clouds without normals by
  parametrizing the gauss formula.
\newblock {\em ACM Transactions on Graphics}, 42(2):1--19, 2022.

\bibitem{xu2023globally}
Rui Xu, Zhiyang Dou, Ningna Wang, Shiqing Xin, Shuangmin Chen, Mingyan Jiang,
  Xiaohu Guo, Wenping Wang, and Changhe Tu.
\newblock Globally consistent normal orientation for point clouds by
  regularizing the winding-number field.
\newblock {\em ACM Transactions on Graphics (TOG)}, 42(4):1--15, 2023.

\bibitem{ji2017surfacenet}
Mengqi Ji, Juergen Gall, Haitian Zheng, Yebin Liu, and Lu~Fang.
\newblock Surfacenet: An end-to-end 3d neural network for multiview stereopsis.
\newblock In {\em Proceedings of the IEEE international conference on computer
  vision}, pages 2307--2315, 2017.

\bibitem{friedman1977algorithm}
Jerome~H Friedman, Jon~Louis Bentley, and Raphael~Ari Finkel.
\newblock An algorithm for finding best matches in logarithmic expected time.
\newblock {\em ACM Transactions on Mathematical Software (TOMS)},
  3(3):209--226, 1977.

\bibitem{rodrigues1840lois}
Olinde Rodrigues.
\newblock Des lois g{\'e}om{\'e}triques qui r{\'e}gissent les d{\'e}placements
  d'un syst{\`e}me solide dans l'espace, et de la variation des coordonn{\'e}es
  provenant de ces d{\'e}placements consid{\'e}r{\'e}s ind{\'e}pendamment des
  causes qui peuvent les produire.
\newblock {\em Journal de math{\'e}matiques pures et appliqu{\'e}es},
  5:380--440, 1840.

\bibitem{rasala1981rodrigues}
Richard Rasala.
\newblock The rodrigues formula and polynomial differential operators.
\newblock {\em Journal of Mathematical Analysis and Applications},
  84(2):443--482, 1981.

\bibitem{zeng2022imagecas}
An~Zeng, Chunbiao Wu, Meiping Huang, Jian Zhuang, Shanshan Bi, Dan Pan, Najeeb
  Ullah, Kaleem~Nawaz Khan, Tianchen Wang, Yiyu Shi, et~al.
\newblock Imagecas: A large-scale dataset and benchmark for coronary artery
  segmentation based on computed tomography angiography images.
\newblock {\em arXiv preprint arXiv:2211.01607}, 2022.

\bibitem{zhang2023multi}
Minghui Zhang, Yangqian Wu, Hanxiao Zhang, Yulei Qin, Hao Zheng, Wen Tang,
  Corey Arnold, Chenhao Pei, Pengxin Yu, Yang Nan, et~al.
\newblock Multi-site, multi-domain airway tree modeling (atm'22): A public
  benchmark for pulmonary airway segmentation.
\newblock {\em arXiv preprint arXiv:2303.05745}, 2023.

\bibitem{maurya2022parse}
Akansh Maurya, Kunal~Dashrath Patil, Rohan Padhy, Kalluri Ramakrishna, and
  Ganapathy Krishnamurthi.
\newblock Parse challenge 2022: Pulmonary arteries segmentation using swin
  u-net transformer (swin unetr) and u-net.
\newblock {\em arXiv preprint arXiv:2208.09636}, 2022.

\bibitem{tetteh2020deepvesselnet}
Giles Tetteh, Velizar Efremov, Nils~D Forkert, Matthias Schneider, Jan
  Kirschke, Bruno Weber, Claus Zimmer, Marie Piraud, and Bjoern~H Menze.
\newblock Deepvesselnet: Vessel segmentation, centerline prediction, and
  bifurcation detection in 3-d angiographic volumes.
\newblock {\em Frontiers in Neuroscience}, page 1285, 2020.

\bibitem{zhang1984fast}
Tongjie~Y Zhang and Ching~Y. Suen.
\newblock A fast parallel algorithm for thinning digital patterns.
\newblock {\em Communications of the ACM}, 27(3):236--239, 1984.

\bibitem{kazhdan2013screened}
Michael Kazhdan and Hugues Hoppe.
\newblock Screened poisson surface reconstruction.
\newblock {\em ACM Transactions on Graphics (ToG)}, 32(3):1--13, 2013.

\end{thebibliography}
\end{document}